\documentclass[10pt,twocolumn,letterpaper]{article}

\usepackage{cvpr}      % To produce the REVIEW version
\usepackage[accsupp]{axessibility}
\usepackage{times}
\usepackage{epsfig}
\usepackage{graphicx}
\usepackage{amsmath}
\usepackage{amssymb}
\usepackage{csquotes}
\usepackage{indentfirst}
\usepackage{algorithm}
\usepackage{algpseudocode}

\usepackage{stfloats}

\usepackage{listings}
\usepackage{xcolor}  % For custom colors
\newcommand{\vc}{{\mathbf{c}}}

\newcommand{\vh}{{\mathbf{h}}}

\newcommand{\vt}{{\mathbf{t}}}

\newcommand{\vx}{{\mathbf{x}}}

\newcommand{\vz}{{\mathbf{z}}}
\newcommand{\vA}{{\mathbf{A}}}

\newcommand{\vP}{{\mathbf{P}}}

\newcommand{\vS}{{\mathbf{S}}}

\newcommand{\RR}{\mathbb{R}}

\newcommand{\ZZ}{\mathbb{Z}}

\DeclareMathOperator*{\argmin}{arg\,min}

\definecolor{cvprblue}{rgb}{0.21,0.49,0.74}
\usepackage[pagebackref,breaklinks,colorlinks,citecolor=cvprblue]{hyperref}

\def\paperID{28} % *** Enter the Paper ID here
\def\confName{CVPR}
\def\confYear{2024}

\title{Context-aware Video Anomaly Detection in Long-Term Datasets}

\author{Zhengye Yang\\
Rensselaer Polytechnic Institute\\
Department of ECSE, Troy, NY, USA\\
{\tt\small yangz15@rpi.edu}
\and
Richard J.~Radke\\
Rensselaer Polytechnic Institute\\
Department of ECSE, Troy, NY, USA\\
{\tt\small rjradke@ecse.rpi.edu}
}

\begin{document}
\maketitle
\begin{abstract}
Video anomaly detection research is generally evaluated on short, isolated benchmark videos only a few minutes long.  However, in real-world environments, security cameras observe the same scene for months or years at a time, and the notion of anomalous behavior critically depends on context, such as the time of day, day of week, or schedule of events.  Here, we propose a context-aware video anomaly detection algorithm, Trinity, specifically targeted to these scenarios.  Trinity is especially well-suited to crowded scenes in which individuals cannot be easily tracked, and anomalies are due to speed, direction, or absence of group motion. Trinity is a contrastive learning framework that aims to learn alignments between context, appearance, and motion, and uses alignment quality to classify videos as normal or anomalous.  We evaluate our algorithm on both conventional benchmarks and a public webcam-based dataset we collected that spans more than three months of activity.
% Video anomaly detection research is generally evaluated on short benchmark datasets only a few minutes long.  However, in real-world environments, security cameras observe the same scene for months or years at a time, and the notion of anomalous behavior critically depends on context, such as the time of day, day of week, or schedule of events.  Here, we propose a context-aware video anomaly detection algorithm, Trinity, specifically targeted to these scenarios.  Trinity is especially well-suited to crowded scenes in which individuals cannot be easily tracked, and anomalies are due to speed, direction, or absence of group motion. Trinity is a contrastive learning framework that aims to learn alignments between context, appearance, and motion, and uses alignment quality to classify videos as normal or anomalous.  We evaluate our algorithm on both conventional benchmarks and a public webcam-based dataset we collected that spans more than three months of activity.
\end{abstract}
    
\section{Introduction}
\label{sec:intro}

\begin{figure*}[!tbp]
\begin{center}
\includegraphics[width=0.99\linewidth]{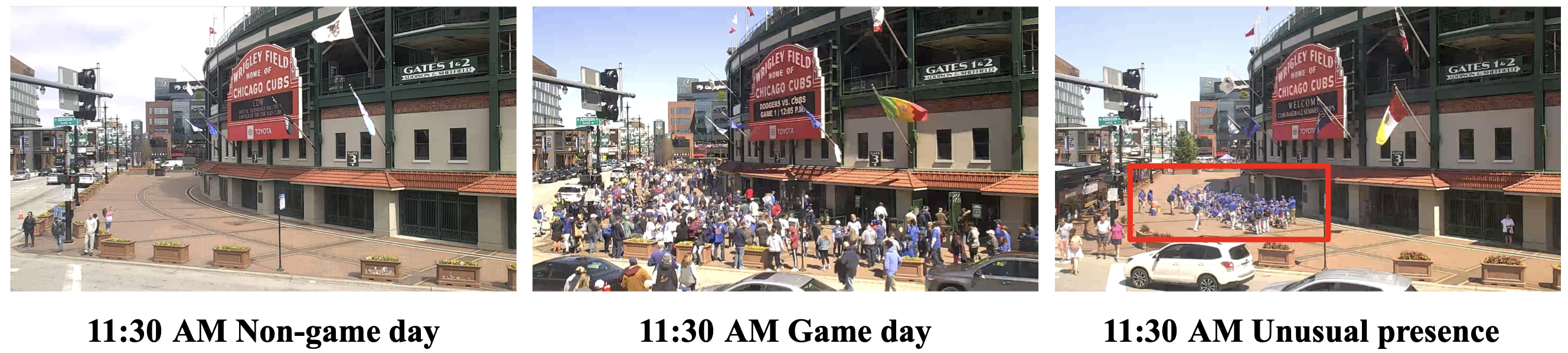} 
\vspace{-0.3cm}
\caption{Snapshots at the same time of day at a baseball stadium. \textbf{Left:} A typical scenario during a non-game day; \textbf{Middle}: A typical scenario during a game day; \textbf{Right}: Unexpected group presence.}
\label{fig:context_anomaly}
\end{center}

\end{figure*}

The goal of video anomaly detection (VAD) is to understand what is ``normal'' and then to identify anomalous events that deviate from the expected distribution. However, the definition of ``normal'' critically depends on \emph{context} information such the time of day, day of week, or schedule of events. For example, a camera monitoring a campus would expect to see groups of people entering campus on a typical weekday morning, but the same scenario during a weekday evening or weekend would be considered anomalous. It would be impossible to accurately detect truly anomalous behaviour without referring to the date and time.  While cameras operating over extended periods have the potential to capture and model such time-varying patterns  \cite{abrams_lost_2012}, current VAD algorithms are generally focused on short, isolated clips and lack any sort of contextual awareness.  This critical gap prevents current algorithms from being immediately applicable to real-world camera networks. 

A na\"ive solution to this challenge is to develop separate VAD models for different contexts~\cite{ramachandra_survey_2020}. However, given that contextual elements like time are continuous, hard classification of contexts can be difficult and inevitably introduce bias. For example, a baseball stadium, as depicted in Figure~\ref{fig:context_anomaly}, may have a varied game schedule, with differing game timings affecting the day's dynamics. Reducing these complex context-dependent dynamics to time-of-day variations by training multiple models is overly simplistic. Creating a context-sensitive VAD framework can aid in detecting anomalies unidentifiable by current techniques, a critical consideration for real-world camera systems, especially those monitoring soft targets such as malls, stadiums, and schools. 

% The normal-anomalous event imbalance, the difficulty of annotation in vast amounts of video data, and the unpredictability of anomalies are major challenges in VAD.  As a consequence of these inherent challenges, previous studies have predominantly approached this problem as either a one-class classification (OCC) \cite{hasan_learning_2016,liu_future_2018,gong_memorizing_2019} or as a weakly supervised task \cite{cho_look_2023,purwanto_dance_2021,sultani_real-world_2018,zhang_exploiting_2022}. This paper primarily approaches the VAD challenge under the OCC setting, which considers all training samples to be normal, and its aim is to detect out-of-distribution samples during inference.  

% Previous work frequently relies on  proxy tasks \cite{xiang_squid_2023,hasan_learning_2016,yang_video_2023,liu_future_2018,georgescu_background-agnostic_2021} to learn the normalcy representation, and these have shown great performance on common benchmark anomaly detection datasets \cite{weixin_li_anomaly_2014,liu_future_2018,lu_abnormal_2013,sultani_real-world_2018}. However, a frequently neglected real-world VAD requirement is the \emph{contextual awareness} of the video's coverage area.  

A few recent VAD datasets recognize the significance of context in anomalies.  Leroux et al.~\cite{leroux_multi-branch_2022} synthetically modified videos (e.g., adding a raindrop overlay) to simulate different conditions to improve the robustness of a VAD algorithm.  The Street Scene dataset \cite{ramachandra_street_2020} highlights spatial anomalies like jaywalking and wrong-way incidents, while the NOLA dataset \cite{doshi_rethinking_2022} collected videos from specific public areas over a week, accompanied by metadata. The latter emphasizes the continual learning aspect of VAD without explicitly correlating video and context. Both temporal and spatial contexts are crucial for comprehensive contextual awareness. To address the long-term contextual awareness requirement for a real-world VAD model, we collected a novel dataset comprising regularly captured videos for a period of three months, augmented with contextual data.

Most VAD benchmarks only contain object-level anomalies (e.g., a cyclist on a pedestrian sidewalk), and do not require holistic scene understanding. While object-centric methods perform highly on standard benchmarks \cite{ionescu_object-centric_2019,georgescu_anomaly_2021,liu_hybrid_2021}, they primarily detect easily-segmented foreground objects. When applied to long-term real-world videos, issues such as complex lighting, diverse weather conditions, and occlusion in crowded settings (as illustrated in Figure \ref{fig:context_anomaly}) make detecting/tracking individuals difficult or impossible, which limits the application of this approach. The computational cost of processing a single clip also scales with the number of objects in a scene. Furthermore, neither object-centric nor frame-based methods are designed to detect absence anomalies (e.g., no one shows up on a school day). 

In this work, our primary goal is to construct a VAD algorithm based on video-context correspondence. We hypothesize that context awareness can be built from the co-occurrence between visual features and their contexts. Inspired by CLIP \cite{radford_learning_2021}, we project visual representations (appearance and motion) and a context representation into a joint embedding space. Appearance and motion can be considered as multiple perspectives of the same video clip, while the context information is an additional modality we want to align with the video clip.

To learn a robust representation in crowded environments and enable contrastive learning of video without context, we consider appearance and motion representations from video clips as multiple views for local alignment.  Through contrastive learning, the context information will guide the model to learn a global representation of the current scenario and detect context anomalies by evaluating the alignment performance. Further, the strong correlation between appearance and motion can enable learning appearance-motion alignment at the local level. During inference, misalignment between appearance and motion can also be used to detect anomalies in traditional VAD benchmark datasets.

The contributions of this paper are as follows:
\begin{itemize}
    \item We propose a real-world VAD task using an exemplar dataset addressing long-term context awareness. 
    \item We propose a new contrastive-learning-based framework called Trinity to learn a joint embedding between modalities globally and locally, which uses alignment error to detect anomalies.   
    \item We present experiments for detecting context-dependent and independent anomalies on both self-collected and benchmark datasets to illustrate the effectiveness of the Trinity framework.
\end{itemize}

\section{Related Work}
\label{sec:related}
\begin{figure*}[tbp]
\begin{center}
\includegraphics[width=0.99\linewidth]{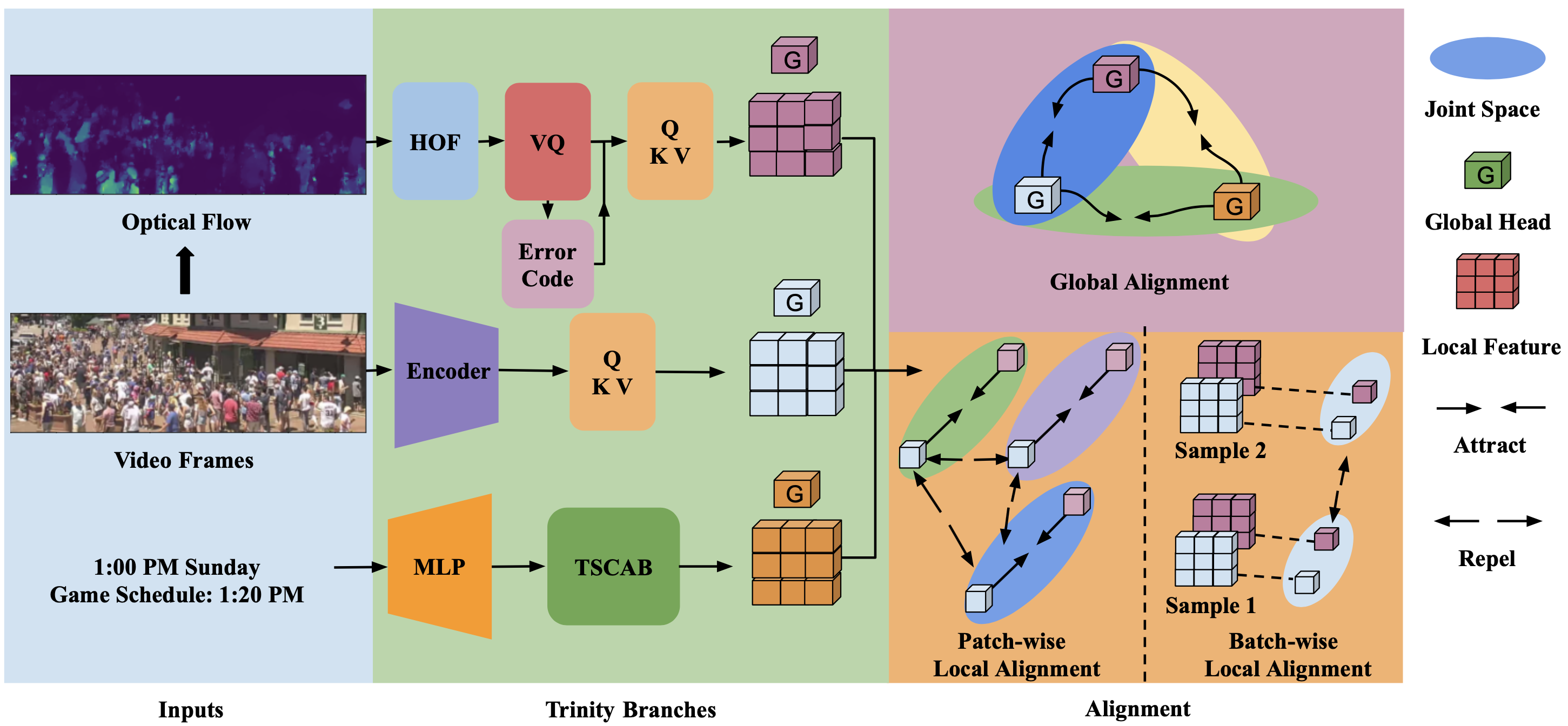}

\end{center}
\vspace{-0.6cm}
\caption{Pipeline for the proposed Trinity algorithm, which takes video frames, optical flow, and contextual information as inputs. The Trinity branches extract global and local representation from each input stream. Global and local alignment are used to learn joint embeddings at different scales and later used to determine anomalies by evaluating misalignments between branches. }
\label{fig:system_overview}
\end{figure*}
\addtocounter{figure}{1}
\textbf{Video anomaly detection.}\ \  Frame-based approaches \cite{zhao_online_2011,hasan_learning_2016,liu_future_2018,gong_memorizing_2019,cho_unsupervised_2022,park_learning_2020,nguyen_anomaly_2019,liu_diversity-measurable_2023,lv_learning_2021} and object-centric approaches \cite{liu_hybrid_2021,ionescu_object-centric_2019,sun_hierarchical_2023,georgescu_anomaly_2021,bao_hierarchical_2022} are the two main branches of detecting video anomalies.
In frame-based methods, detecting anomalies is typically based on reconstruction error. For example, Hasan \etal~\cite{hasan_learning_2016} proposed to use a convolutional autoencoder (AE) in which reconstruction failures indicate anomalies. Gong \etal~\cite{gong_memorizing_2019} proposed MemAE, using an external memory array that is jointly optimized with an AE to further enlarge the reconstruction error between normal and abnormal objects. Another line of work also takes an AE as its main component to predict the next frame and uses the prediction discrepancy as the anomaly indicator \cite{liu_future_2018,lv_learning_2021,ristea_self-supervised_2022,park_learning_2020}. MNAD \cite{park_learning_2020} concatenated encoded features with the closest prototypical pattern to pass onto the decoder. Recently, DMAD \cite{liu_diversity-measurable_2023} added a deformable model to capture nuances between normal variations and anomalous events to improve performance. The recent progress of unsupervised VAD has slowly shifted to object-centric methods \cite{ionescu_object-centric_2019,georgescu_anomaly_2021,liu_hybrid_2021,sun_hierarchical_2023}, which use pre-trained object detection or tracking algorithms to segment objects used as the input instead of the whole frame. HF$^{2}$VAD \cite{liu_hybrid_2021} proposed object detection and a multi-level memory module with shortcuts to memorize prototypical flow patterns and used its reconstruction to further condition on the next frame prediction. These newly proposed methods use the strong prior that only moving objects are considered as anomalous region candidates, which cannot detect absence/presence type anomalies. 

% \rnote{what?} 
\textbf{Context modeling.}\ \  Prior work on building scene awareness into VAD \cite{bao_hierarchical_2022,sun_hierarchical_2023} can be considered as capturing a special case of context. For example, Sun~\etal~\cite{sun_scene-aware_2020} proposed a spatial-temporal context (STC) graphical model to capture context information about the current video tube to improve the performance of VAD. It treated bounding boxes from object detection as nodes for building the context graphic model. Sun and Gong \cite{sun_hierarchical_2023} used scene segmentation, tracklets, and a pretrained vision transformer \cite{dosovitskiy_image_2021} to parse the video.  They used memory modules to memorize the prototypical patterns of scene-appearance and scene-motion features with contrastive learning to enlarge the difference between different scene representations.  
% \rnote{Talked about this paper in the previous paragraph, you only should mention it once.}
% Compared to building a scene-awareness, formulating long term context is a more challenging term since the background information is more distinctive and the background remain constant for each camera in benchmark datasets. 
% Most scene-aware methods concatenate the background representation with object features to formulate scene-specific representations.  This might cause redundancy in the memory due to concatenation. \rnote{??} 

Context-aware anomaly detection in more general modalities has been studied \cite{song_conditional_2007,liang_robust_2016,li_explainable_2023}. For example, Song \etal\cite{song_conditional_2007} proposed a method called Context Anomaly Detection (CAD) that uses a Gaussian mixture model to model context feature and behavior feature distributions separately with regression mapping as a linkage to detect context anomalies. As mentioned in Tian \etal \cite{vedaldi_contrastive_2020}, a predictive objective is less effective than using a contrastive objective for learning a better representation under the multi-view setting. In this paper, we use contrastive learning to align the context information and visual representations to build contextual awareness without using auxiliary pre-trained models for context modeling.  
% and Quantile-based CAD (QCAD)~\cite{li_explainable_2023} utilizes quantiled regression forest to perform context anomaly detection.

% Previous work~\cite{leroux_multi-branch_2022} applies augmentation on benchmark data to simulate the lighting variation and weather changes. However, the lighting and weather changes not only modify the background appearance but should also affect the foreground objects' behaviors (People don't walk the same as usual when having a downpour).

\textbf{Long term datasets.}\ \ The LOST dataset \cite{abrams_lost_2012} provided videos from 19 cameras from the same half-hour each day over the course of a year, along with metadata. Through trajectory clustering, the aggregated statistics show strong correlations with external metadata, but most of the following work focused on trajectory prediction or trajectory abnormality detection \cite{dotti_unsupervised_2017,xie_learning_2018} without investigating the context anomalies. The NOLA dataset \cite{doshi_rethinking_2022}  consistently collected videos in specific public areas over a week, accompanied by metadata. Although it contains context-dependent anomalies like loitering, the random camera position change makes it unsuitable for frame-based methods. It emphasizes the continual learning aspect of VAD without explicitly correlating video and context. The LTD dataset \cite{nikolov_seasons_nodate}  collected 8 months of videos from a thermal camera with context information like weather and time for investigating concept drifts like seasonal changes. In this paper, to better investigate the video-context correspondence, we collect a new dataset containing videos with context information targeting a baseball stadium that has a complex game schedule and potentially crowded scenarios. 

\section{Method}
\label{sec:method}

% \rnote{Center the tiny G blocks above the 3x3 arrays in the 2nd panel.  Not clear on how the 3rd panel relates to the first two (no input/output?}
% To achieve unsupervised VAD with ability of building context-video correspondence, CAMVAD takes video frames and provided temporal contexts as input, train the model to memorize the background and foreground object and inject contextual information during memory gating mechanism. 

The overview of our proposed method, which we call Trinity, is shown in Fig.~\ref{fig:system_overview}. Trinity contains three branches: a context branch, an appearance branch, and a motion branch. In this section, we first introduce the design of each branch and describe the training scheme of the model. We then discuss how to use global and local alignment to detect context anomalies.  

\subsection{Branches of the Trinity Model}  
Consider a batch of video clips $ \{ \vx_i,\ldots, \vx_B \} \in \RR^{ C \times T\times H \times W} $, where $B,C,T,H,W$ denote batch size, channel size, number of frames,  height, and width respectively.  For later processing, each frame is split into a $k \times k$ grid of nonoverlapping patches.
The corresponding context information is $\{\vc_1,\ldots,\vc_B\}  \in \ZZ^{Y}$, where $Y$ represents the dimension of concatenated  one-hot encodings of context information, discussed further below.  Each branch takes different source information as input but generates the same-sized representation as output, containing the concatenation of a global head token and a local token resulting in $\vh  \in \RR^{B \times N \times D }$, where $N = \frac{H\times W}{k^2} +1$ and $D$ represents the dimension of the joint embedding space. Detailed structure information is provided in the Supplementary Material \cref{sec:Supp_archtecture}. 

\subsubsection{Context Branch}
The context information is treated as a concatenation of one-hot encodings of each type of context. For example, in our stadium dataset, different types of context include the hour of day, day of week, a flag for game/non-game day and the corresponding game schedule.  We use a simple two-layer MLP to project the context information into a context embedding: 
$\{\vP_{1}^{\vt},\ldots, \vP_{B}^{\vt}\} \in \RR^{d_t}$. To enable the local alignment between context and other modalities, we use a learnable positional encoding: 
$\vP^{\mathbf{s}}_{i} \in \RR^{d_p}, i \in \{1,\ldots,N\}$ embedding as the context prototypical representation.  

\textbf{Temporal/Spatial Context Attention Block.}\ \  To learn the relationship between temporal and spatial context in an unsupervised fashion, we propose a Temporal/Spatial Context Attention Block (TSCAB) that is able to adjust the spatial embedding $\vP^{\mathbf{s}} $ based on contextual information $\vP_{j}^{\vt}$ at batch index $j$. The key ingredient here is to intertwine information between the positional embedding and the temporal context embedding through Multi-head Cross Attention (MCA) \cite{vaswani_attention_2023} as below: 
\begin{equation}
% \begin{split}
\begin{aligned}
\hat{\vP}^{\mathbf{s}}_{i} &=W_{proj} \cdot \text{MultiHead}  \\
\text{MultiHead} &= Softmax\left(\frac{Q\cdot K^T}{\sqrt{d_p}}\right)\cdot V  \\
Q = W_Q\cdot \vP_{j}^{\vt},\ \ K &= W_K\cdot \vP^{\mathbf{s}}_{i},\ \ V = W_V\cdot \vP^{\mathbf{s}}_{i} 
% \end{split}
\end{aligned}
\label{eq:cross}
\end{equation}
Through a series of TSCAB operations, we obtain the augmented positional embedding $\hat{\vP}^{\mathbf{s}}_{i}$ that serves as the basis to align features from other branches via a projection $W_{proj}^{cxt} \in \RR^{ d_p \times D }$ shown in (\ref{eq:pe_proj}), where $[\cdot]$ is the concatenation operation. The context representation is $\vh_{cxt}  \in \RR^{B \times N \times D }$. 

\begin{equation}
\vh_{cxt} = [\hat{\vP}^{\mathbf{s}}_{1}, \ldots,\hat{\vP}^{\mathbf{s}}_{N}] \cdot  W_{proj}^{cxt}
\label{eq:pe_proj}
\end{equation}

\subsubsection{Motion Branch}

\textbf{Feature preparation.}\ \  We use the Histogram of Optical Flow (HOF) as the representation of motion for its simplicity. We use the TV-L1 method \cite{hamprecht_duality_2007} to extract a pixel-wise flow field and divide the input into non-overlapping patches for calculating the histogram of flow. Following \cite{singh_eval_2022}, each patch contains 12 orientation bins, a background ratio representing the proportion of pixels with magnitude below a threshold, and the average magnitude for each orientation bin, resulting in a total of 25 channels for each local patch. 

\textbf{Vector Quantization.}\ \ In our initial attempts, we found that using a convolution block to extract flow information tends to downplay the importance of flow magnitude. To better preserve the magnitude information, we use a vector quantization layer \cite{van_den_oord_neural_2017} to formulate a vocabulary $\vz_n \in \RR^{25}, n \in \{1,\ldots, M\} $ with vocabulary size $M$ after extracting the HOF features $\vz_{mot}^{e}$ shown in (\ref{eq:vq}). The index of the quantization vector is used as the input token to the following transformer \cite{vaswani_attention_2023} blocks. 
\begin{equation}
    \vz_{mot}^{q} = \argmin_{\vz_n}||\vz_{mot}^{e}  - \vz_n ||_2
\label{eq:vq}
\end{equation}

\textbf{Error Code Usage.}\ \ Using the vector quantization layer for tokenizing motion has one problem: it is unable to pass unseen motions into the motion branch. To identify unseen motions, we propose a simple method called Error Code Usage. For the given HOF feature $\vz_{mot}^{e}$, we obtain the quantized feature $\vz_{mot}^{q}$, and further encode the residual between  $\vz_{mot}^{e}$ and $\vz_{mot}^{q}$ using the same vocabulary to get the error word $\vz_{mot}^{err}$. The intuition behind the error code is normally that the VQ layer can find a $\vz_{mot}^{q}$ that is close to $\vz_{mot}^{e}$, leaving a residual that is nearly the zero vector. The following model will learn this as a normal word pair. If an unseen motion happens, the residual word contains a large motion and the resulting word pair is unseen, causing misalignment between the branches. 

We use the same embedding to project both $\vz_{mot}^{q}$ and $\vz_{mot}^{err}$ and concatenate the representation, feeding word pairs of the whole clip into $k_{mot}$ transformer blocks. After the transformer blocks, we use a simple projection to generate the motion representation $\vh_{mot}  \in \RR^{B \times N \times D }$ for joint embedding alignment. 

\subsubsection{Appearance Branch}

Since Trinity is agnostic to the appearance feature extraction model, we simply adopt the basic U-net structure from MNAD \cite{park_learning_2020}, which is widely used in the VAD community. The U-net framework contains an encoder $f(\vx)$ and a decoder $g(\cdot)$ with shortcuts in between to perform the reconstruction task shown in (\ref{eq:recon_loss}). 
\begin{equation}
     L_{recon} = || \vx - g(f(\vx))||_2
\label{eq:recon_loss}
\end{equation}
The latent representation $\vz$ is extracted through the encoder $f(\vx), \vz \in \RR^{d' \times  h' \times w'}$. To avoid adding disturbance to the reconstruction, we apply the stop gradient to the extracted features before feeding into the transformer blocks.
To tokenize the extracted feature map, we use a 2D convolution layer with the same kernel size and stride $(r_h,r_w) = (\frac{h'}{h},\frac{w'}{w})$ to embed features from the deepest layer of the encoder. Similar to the motion branch, after $k_{app}$ transformer blocks and a final projection layer, we obtain the appearance representation $\vh_{app}  \in \RR^{B \times N \times D }$.

The three Trinity branches result in $\vh_{app}$, $\vh_{mot}$, and $\vh_{cxt}$, the output representations for the alignment training described in the next section.

\subsection{Aligning Feature Representations}
 The following alignment process contains both global and local alignment. The global representation is denoted as $\vh^{g} \in \RR^{B \times 1 \times D }$, and the local representation is denoted as $\vh^{l} \in \RR^{B \times N-1 \times D }$, where $\vh = [\vh^{g},\vh^{l}]$.   Our focus is on context-dependent anomalies; in Section 4.5 we describe how the framework can be modified to also detect context-free anomalies (i.e., the traditional VAD problem).

\subsubsection{Global Alignment}
The global representation gathers the information from local patches through transformer blocks to provide a holistic representation of the current video clip. This is vital for formulating context awareness. Global alignment treats all samples within the batch except the target sample as negative samples. We use the extracted global tokens from selected branch pairs $\vh^{g}_{1}, \vh^{g}_{2}$ as inputs.
% and calculate the cosine similarity matrix $\vA$ of samples between normalized global representations. 
The global alignment loss for a pair $L^{global}_{1,2}$ is calculated through the symmetric binary cross entropy loss as employed in CLIP \cite{radford_learning_2021} shown in (\ref{eq:global_bce}), where $+$ is the indicator of paired sample and $\tau$ is a learnable temperature parameter. 
\begin{gather}
L^{global}_{1,2} = -\sum_{i=1}^{B} \left( \log \text{NCE}(\vh^{g}_{1},\vh^{g}_{2}) +\log \text{NCE}(\vh^{g}_{2},\vh^{g}_{1})\right) \notag\\
\text{NCE}(\vh^{g}_{1},\vh^{g}_{2})  = \frac{\exp(\vh^{g}_{1}\cdot (\vh^{g}_{2})^{+}/\tau)}{\sum_{i=1}^{B} \exp(\vh^{g}_{1}\cdot \vh^{g}_{2}/\tau)} 
\label{eq:global_bce}
\end{gather}
For context video anomaly detection, the motion-context and appearance-context pairs are selected to build the global alignment resulting in $L_{global} = L^{global}_{app, cxt}+ L^{global}_{mot, cxt}$. 
% \rnote{missing symbol??}
% \rnote{edge ???}
\subsubsection{Local Alignment}

At the local level, the reference negative samples can be separated into two categories: batch-wise samples and patch-wise samples. These two categories address different aspects of information alignment by applying different negative sampling strategies.

\textbf{Batch-wise local alignment} treats all the local patches at the same location across a batch as negative samples.  It tries to align the information from two selected branches $\vh^{l}_{1}, \vh^{l}_{2}$ at the local level, which is similar to the global alignment. After getting similarity logits with shape $(N\times B\times B)$, the batch-wise local alignment loss $L^{batch}_{1,2}$  is calculated similarly as $L^{global}_{1,2}$ with average $N$ batch alignment loss performed locally. This type of alignment is not suitable for cross-modality alignment (e.g., context-motion alignment) but can be applied between multiple views (e.g., motion-appearance alignment) since cross-modality information might not have a consistent correlation at the local level. In the later ablation study, we further validate that cross-modality local alignment causes a performance drop. 

\textbf{Patch-wise local alignment} treats the rest of the local patches within a video clip as negative samples. It aims to articulate the spatial context, pushing the feature representations from different positions away from each other. The patch-wise local alignment loss $L^{patch}_{1,2}$ is calculated by averaging the $B$ alignment results using the same symmetric binary cross entropy loss of logits with shape ($B\times N\times N$).  For context video anomaly detection, the local level alignment loss $L_{local}$ is:
% \rnote{2nd and 4th terms are the same.  Notation is confusing.}
\begin{equation}
L_{local} = L^{batch}_{app,mot} + L^{patch}_{cxt,mot} +  L^{patch}_{cxt,app} + L^{patch}_{app,mot}
\label{eq:cxt_local}
\end{equation}

\subsection{Training and Inference}
\textbf{Training.}\ \ We pretrain the vector quantization layer before training the full Trinity model using the simple MSE loss. Given a video with provided context, the training loss combines the reconstruction loss, the global alignment loss, and the local alignment loss with weighing parameters:
\begin{equation}
L =  L_{recon} + \beta_1 \cdot L_{local}+ \beta_2 \cdot L_{global} 
\label{eq:full_loss}
\end{equation}

\textbf{Inference.} \ \ During inference, context anomalies can be detected by evaluating the global context-motion similarity and context-appearance similarity. The global context similarity $\vS_g$ is the equally weighted average of the alignment logits after applying the sigmoid function $\sigma$: 
\begin{equation}
\vS_g = \frac{1}{2}\left(\sigma\left(\frac{\vh_{cxt} \cdot \vh_{mot}^T}{\tau}\right)+\sigma\left(\frac{\vh_{cxt} \cdot \vh_{app}^T}{\tau}\right)\right)
\label{eq:global_cxt}
\end{equation}
We combine this with the widely adopted next frame prediction quality from the U-net using the min-max normalized PSNR $\vS_r$ with a weighing parameter $\alpha$ for detecting non-context anomalies. We compute the  final normalcy score as $\vS = \alpha \cdot \vS_r+ (1-\alpha)\cdot \vS_g$ (and in practice graph the anomaly score $1-\vS$).

% \rnote{I don't come away from this with a clear understanding of conceretely how the anomaly decision is made.  Also not clear on what non-context related anomalies are here.} 

% we adopt  in (\ref{eq:psnr}) as the indicator of normalcy.   
% \begin{equation}
% \text{PSNR}(\vx,\Tilde{\vx}) =10\log_{10}\frac{[\max{\Tilde{\vx}}]^2}{\frac{1}{N}\sum_{i=0}^{N}( \vx-\Tilde{\vx})^2}
% \label{eq:psnr}
% \end{equation}

\section{Experiments and Analysis}
\label{sec:exp}

% \begin{table*}[!t]
%  \begin{center}
%     {
% \begin{tabular}{c|c| c|c|c|c}
% \toprule
% \textbf{Dataset} & \textbf{Hours} & \textbf{Anomaly events} & \textbf{Anomaly type} & \textbf{Context Anomaly} & \textbf{Ground Truth}  \\
% \midrule
% UCSD Ped2~\cite{weixin_li_anomaly_2014} & 0.4 & 20 & 5 &context-free & pixel masks \\
% Avenue~\cite{lu_abnormal_2013} & 0.5 & 47 &  5 &  context-free& pixel masks \\
% ShanghaiTech~\cite{liu_future_2018} & 3.6 & 130 & 17  & context-free & pixel masks \\
% Streetscene~\cite{ramachandra_street_2020} & 4 & 205 & 17  &spatial  & bounding boxes \\
% ADOC~\cite{ishikawa_day_2021} & 24 & 721 & 25 & spatial  & bounding boxes \\
% NOLA~\cite{doshi_rethinking_2022} & 1.3& 40 & - & context-free  & time stamp \\
% \midrule
% \textbf{Ours} & \textbf{32} & 64 & 9 & \textbf{spatial+temporal} & Tracks \\
% \bottomrule
% \end{tabular}
% }
%  \end{center}
% \caption{Video Anomaly Detection Dataset characteristics \rnote{Where is this table referenced?  While nice to have it takes up space we could be using for something else. What does it mean that our ground truth is ``tracks''?}}
% \label{tab:dataset_stat}
% \end{table*}
% \TODO{shrink text and move table stats into text description}

\subsection{The WF Dataset}\label{sec:dataset}

Benchmark VAD datasets mainly target anomalies that solely depend on individual objects and don't consider temporal context.  We therefore collected a new dataset, denoted as the WF dataset, from a public webcam looking at a baseball stadium to address the video-context correspondence problem.  We collected 2000 1920 $\times$ 1080 resolution videos of approximately 1 minute length at 15 frames per second across a period of three months.  The collection process has a 30-minute gap between each run, resulting in a large set of videos of the same location under various lighting, weather, and crowding conditions.  We also collected corresponding metadata including the time of the day, day of week, and scheduled game time.  To address the potentially populated region of interest, we cropped a 400$\times$1200 region out of the original frames for both training and testing. More details can be found in the Supplementary Materials \cref{sec:supp_data}.  

% Current benchmark datasets are collected under well-lit environments, which only represents a small fraction of daily scenario. Various lighting and weather environments are crucial for deploying VAD in real life.  
\textbf{Anomalies in the wild.}\ \ The anomalous behaviors in the WF dataset are a mixture of context-dependent anomalies (i.e., behaviors whose anomalousness depends on time of day, game schedule, etc.) and context-free anomalies (i.e., behaviors that would be anomalous regardless of when they occurred).  Context-dependent anomalies we found included people running in the middle of the night, an ambulance driving against normal traffic flow, and unexpected group presence in front of the stadium.  This resulted in 12 videos containing 64 anomalous events with bounding box annotations, which we call True Anomalies.  
% \rnote{This is confusing, since in the next section we say we only have 12 true anomalies.  Also the distinction between context and context-free anomalies is unclear (especially after we made a big deal about it).} 

\begin{figure}[!thb]
\begin{center}
\includegraphics[width=0.99\linewidth]{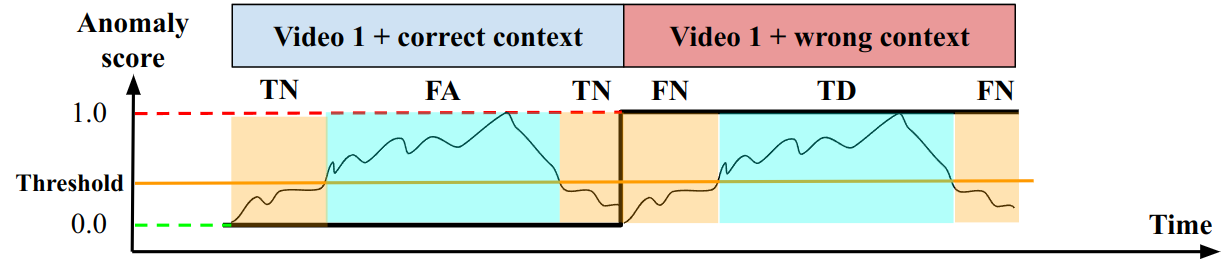}
\caption{Example of pseudo context anomaly evaluation. The algorithm is expected to identify whether the given context matches the input video. TN, TD, FA, FN are true negative, true detection, false alarm and false negative respectively.\vspace{-.75cm}}
\label{fig:rebuttal_1}
\end{center}
\end{figure}

\textbf{Pseudo-contextual anomalies.}\ \ A well-known problem in real-world VAD is the lack of natural anomalous events in proportion to  normal observations.  To compensate, we introduced what we term pseudo-contextual anomalies, which are simply existing videos in the dataset given an incorrect context label. For example, a video recorded just before a game starts should contain frames of people entering quickly into the stadium.  If we modify the context indicating the game has already started two hours ago, the model should flag the whole video as anomalous. Following this approach, we selected 20 normal videos covering various contexts. During testing, each video is evaluated twice: once with the original context and again with an incorrect context.  The ideal result is that all the frames in the first case are normal and all the frames in the second case are anomalous.   In contrast, as Fig.~\ref{fig:rebuttal_1} illustrates, a context-unaware algorithm will have a 50.0 AUC score in this evaluation (since it has the same ratio of positive detections to false alarms for each double trial). 
 % We select several examples shown in Fig.~\ref{fig:pseudo_aomaly}.
\subsection{Benchmark Datasets and Evaluation Metrics}

In addition to the new dataset we introduced above, we also evaluate the model performance in the widely used benchmark datasets UCSD Ped2 \cite{weixin_li_anomaly_2014}, Avenue \cite{lu_abnormal_2013} and ShanghaiTech \cite{liu_future_2018}. All three benchmarks treat walking pedestrians as normal events, while other objects like bikers and abnormal motions like dropping bags or running are considered as anomalies.  ShanghaiTech contains 13 different scenes, while Ped2 and Avenue are single scene datasets.

We created a context-augmented dataset called the Biker Day dataset based on the Ped2 dataset to verify the effectiveness of our context-dependent approach. The dataset is created by adding a context called Biker Day that considers both pedestrians and bikers as normal objects. To create this dataset, we included test videos [Test001, Test002, Test003] that contain bikers from the Ped2 test set into the training data and removed these  videos during evaluation.  If we turn off the Biker Day context, the bikers will again become anomalous as in the original Ped2 dataset. Since all selected videos also contain frames in which bikers are not present, those frames should be considered normal on a Biker Day.

We use the widely adopted area under the curve (AUC) of the receiver operating characteristics (ROC) as the frame-level metric to report the model performance in the following experiments.
% \rnote{Should also say why you chose the comparison algorithms that you did.  Again, the emphasis should be that other algorithms are not made for our dataset, and conversely that we don't care that much about solving the problem posed by standard benchmarks.}

\subsection{Implementation Details}

For Ped2, Avenue, and ShanghaiTech, all frames are resized to 256$\times$256 and the frame sampling gap is 1.  For the WF dataset, all frames are resized to 128$\times$384 and the sampling frame gap is 4. The patch size for all videos is set to 16. The RGB values of all frames are normalized to [-1,1]. The length of the input frame tube is 4. We use 1 and 2 TSCAB blocks in the Biker Day and WF datasets respectively. The size of the vector quantization codebook is 512. The initial temperature $\tau$ is set to 0.07. $\beta_1$ and $\beta_2$ are set to 1.0. The learning rate used for the pretraining procedure of the vector quantization layer is 1e-3. The model is optimized by Adam \cite{kingma_adam_2017} and the learning rate is 2e-4 with the cosine annealing \cite{loshchilov_sgdr_2017} strategy. The total training epochs for Ped2, Avenue, ShanghaiTech, and WF are 60, 60, 10, and 20 respectively. The weighting parameter $\alpha$ for WF dataset and benchmark datasets is 0.3 and 0.7 respectively. We adopt a simple median filter to smooth the normalcy result with a kernel size of 17 following \cite{liu_hybrid_2021}. We provide more details in Supplementary materials \cref{sec:supp_data}.

\subsection{Detecting Context Anomalies}

\begin{figure*}[b]
\begin{center}
\includegraphics[width=0.99\linewidth]{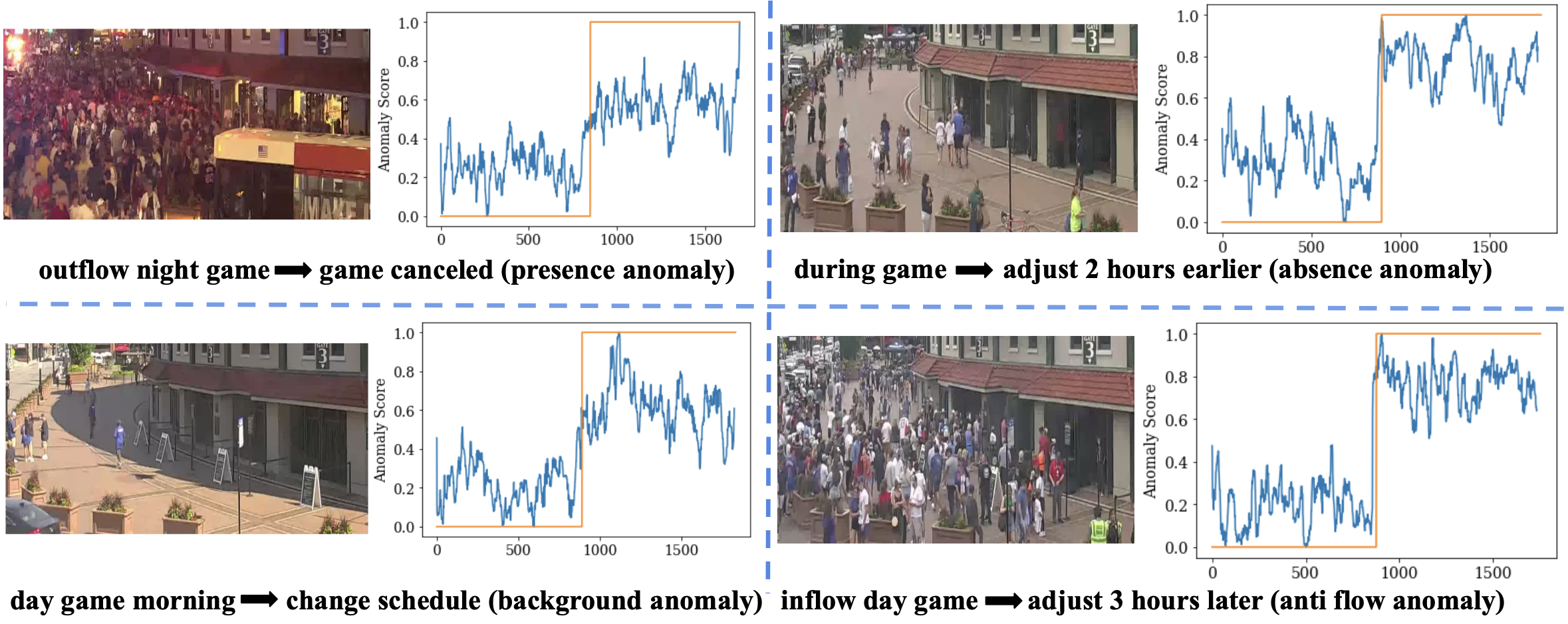}
\vspace{-0.7cm}
\end{center}
\caption{Selected results of pseudo anomalies in the WF dataset. \textbf{Top:} Selected pseudo anomaly detection results with exemplar frames. The orange line indicates ground truth and the blue line is corresponding prediction. The caption indicates the change from the true context (left of the arrow) to the pseudo context (right of the arrow). The parentheses indicate the type of context anomaly. }
\label{fig:pseudo_aomaly}
\end{figure*}
\addtocounter{figure}{-2}

\begin{figure}[thbp!]
\begin{center}
\includegraphics[width=0.99\linewidth]{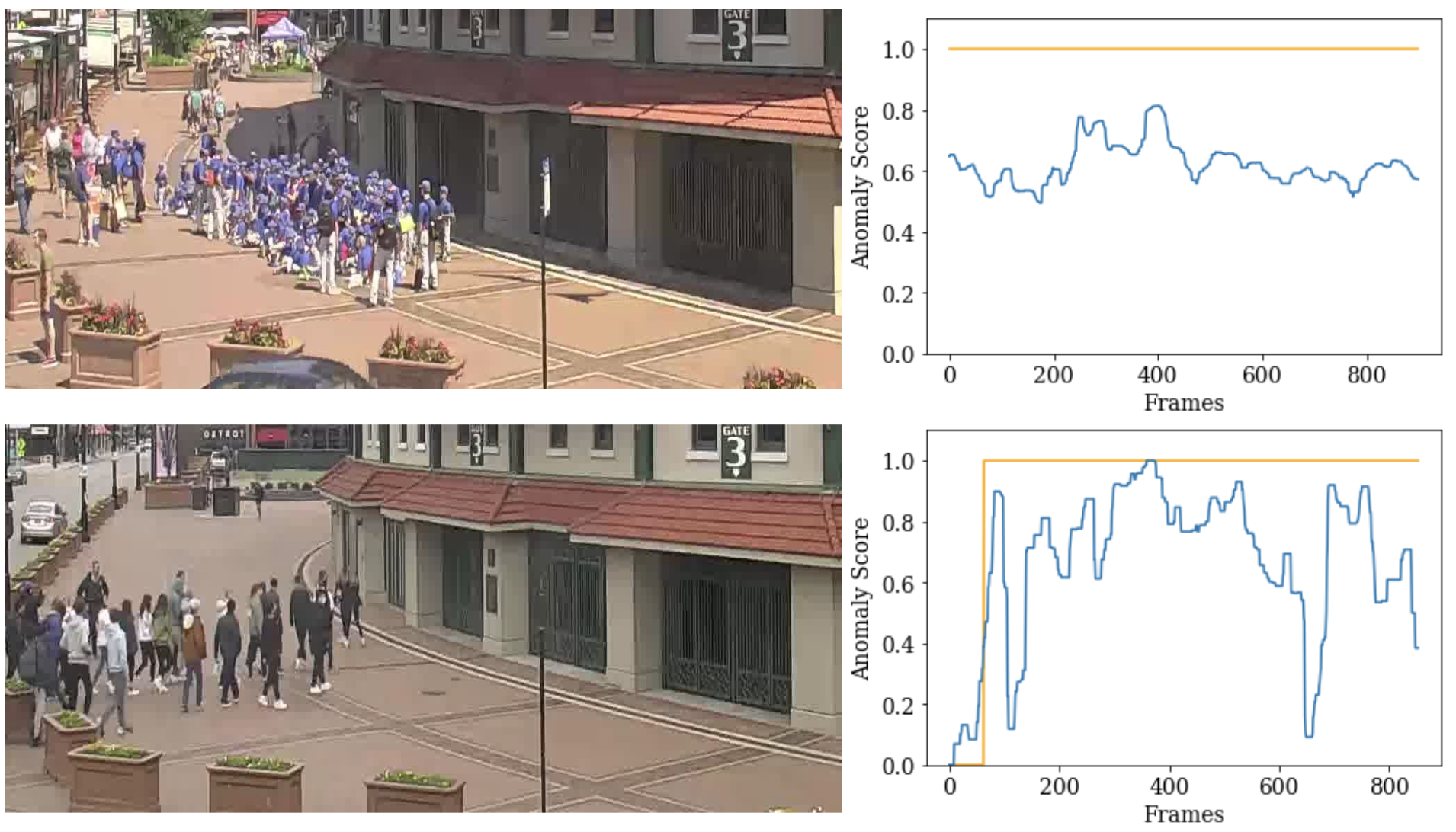}
\vspace{-0.3cm}
\caption{Real context anomalies in the WF dataset. These two sample videos contain unexpected crowds at the front of the stadium (presence anomalies). The blue line is the prediction result and the orange line is the ground truth. }
\vspace{-0.4cm}
\label{fig:real_anomaly}
\end{center}
\end{figure}

\textbf{WF Dataset True Anomalies.}\ \  We applied several widely-used VAD algorithms (previously evaluated on common benchmarks) to our WF dataset to compare to our proposed method. We use officially released codebases and follow all the hyperparameter settings used in the original implementations except the resolution. The evaluation performance is shown in Table~\ref{tab:wf_auc}. These referenced algorithms' overall performance on the real anomalies shows relatively similar ranking as in benchmark datasets.  Since the WF real anomalies contain a significant portion of context-free anomalies, the referenced algorithms can perform well on detecting them.  However, they perform poorly at detecting out-of-place presence/absence/motion that only seem unusual in context.  With the help of global context alignment, Trinity is able to achieve the best performance with a noticeable performance margin. We visualize two 
testing samples containing real context anomalies with the prediction score using Trinity shown in Fig.~\ref{fig:real_anomaly}. The global alignment is quite robust in detecting contextual anomalies. 

\begin{table}
 \begin{center}
    {
\begin{tabular}{c|c|c}
\textbf{Method}  & \textbf{Real Anomaly }& \textbf{Pseudo Anomaly}   \\
\hline
ConvAE \cite{hasan_learning_2016} & 69.4& 50.0 \\
MemAE \cite{gong_memorizing_2019} & 71.3& 50.0 \\
MNAD \cite{park_learning_2020}  & 72.3&  50.0 \\
\hline
\textbf{Ours}  & \textbf{75.5} & \textbf{92.0} \\
\end{tabular}\vspace{-.8cm}
}
 \end{center}
\caption{Contextual anomaly detection results on the WF dataset.\vspace{-.6cm}}
\label{tab:wf_auc}
\end{table}

\textbf{WF Dataset Pseudo Anomalies.}\ \ Out-of-context detection is the main purpose of the pseudo anomaly test. However, previous VAD methods are not equipped with the ability to detect out-of-context anomalies, finding each double-length video to be completely normal.  On the other hand, Trinity performs extremely well in detecting out-of-context anomalies such as presence, absence, and counter-flow direction, with an AUC of 92.0.  We illustrate several pseudo anomaly detection results in Fig. \ref{fig:pseudo_aomaly}. These types of anomalies are impossible to detect using traditional VAD methods. Here, the context-appearance and context-motion branches complement each other very well. If only use a single branch, the pseudo anomaly performance is 76.5 and 87.9 AUC for the context-appearance and context-motion branch respectively. This is because different context anomalies depend on different types of information. For example, the only cue for detecting an anomaly in game schedule (lower left of Fig.~\ref{fig:pseudo_aomaly}) are some large sandwich boards set out on the pavement ``too early'', which cannot be detected by motion information. Similarly, for the anti-flow anomaly introduced at the lower right of Fig.~\ref{fig:pseudo_aomaly}, it is not easy to distinguish whether the crowd is entering or exiting the stadium from only the static image. 

\addtocounter{figure}{1}

\begin{figure}[!h]
\begin{center}
\includegraphics[width=0.99\linewidth]{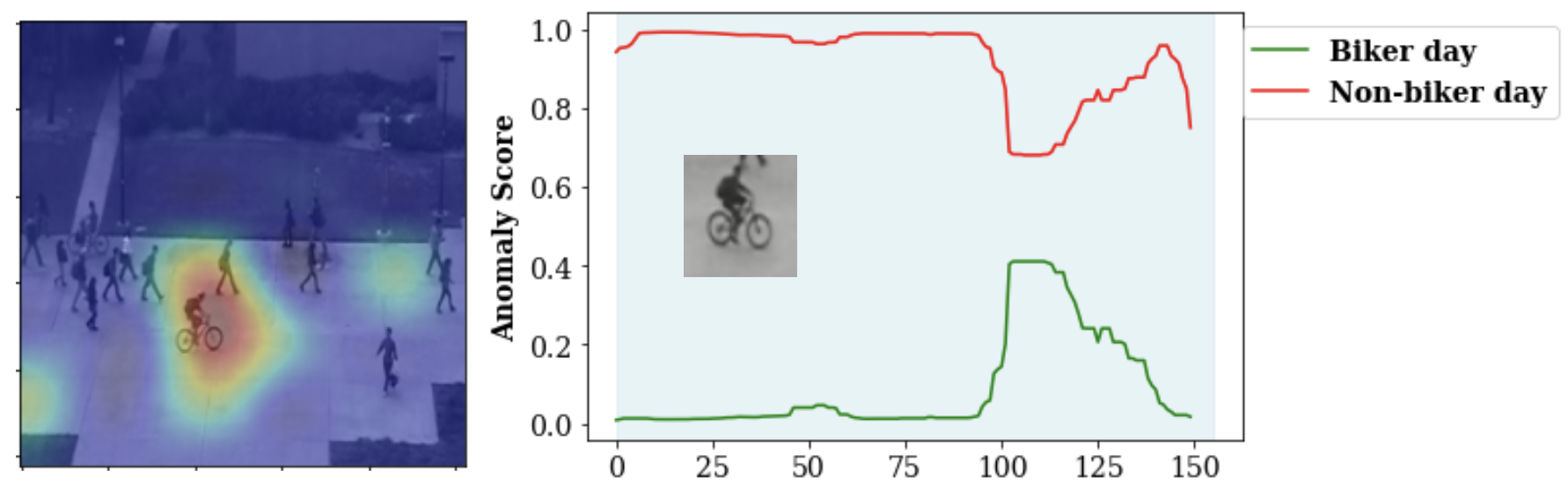}
\vspace{-0.3cm}
\caption{\textbf{Left:} Visualization of global context-appearance head attention on the Biker Day dataset. Biker regions have been assigned highest attentions across all local patches. The attention map has been enlarged for image overlay. \textbf{Right:} The anomaly score under Biker Day and Non-Biker Day context on `Test Video010' from Ped2. The biker region has been highlighted using light blue.}
\label{fig:biker_att}
\end{center}
 \vspace{-0.4cm}
\end{figure}

\textbf{Biker Day experiment.}\ \ Without Trinity global alignment, solely using the U-net from Section 3.1.3 on the augmented dataset resulted in 92.4 AUC, compared to 94.3 in the original Ped2 dataset, which is a significant drop due to the mixture dataset containing bikers as normal objects. By learning the Biker Day context, the global alignment can differentiate the context and boost the performance from 97.9 in Ped2 to \textbf{98.5} AUC. To further validate that the Trinity model really learns the Biker Day context, we visualize the attention map for the global head of the context-appearance pair and the context-controlled anomaly detection result of `test video010' from the Ped2 dataset in Fig.~\ref{fig:biker_att}.  We see that Trinity highlights the biker region with high attention to differentiate context. The `test video010' only contains a biker as an anomaly across the entire video. By switching the input context, Trinity assigns a high anomaly score under the Non-Biker Day context and a low anomaly score under the Biker Day scenario. 

% \rnote{I don't think this really sells the experiment.}

% \rnote{I don't understand what we're looking at here or why it illustrates the Biker Day concept well.}

% \rnote{Needs more interpretation to sell the point; this is the key result!}

\textbf{Ablation studies.}\ \ We perform ablation studies on the impact of different combinations of alignment schemes and the importance of vector quantization, summarized in Table \ref{tab:WF_ablation}. The vector quantization and patch-wise local alignment improve the context detection performance significantly while adding the batch-wise local alignment causes the performance drop. This indicates that the context information of the video does not have strong mutual information with motion at the local level, which suggests that the cross-modality alignment should not focus on local precise alignment but on finding a good global representation.  

\begin{table}[hbtp!]
 \begin{center}
    {
\begin{tabular}{c c c c | c c}
\textbf{GA} & \textbf{VQ}& \textbf{LB} &  \textbf{LP} &  \textbf{AUC$_{mot}$} & \textbf{AUC} \\
\hline
\checkmark & &  & & 83.8 & 89.3\\\checkmark & \checkmark&  & & 86.4 & 91.2\\
\checkmark & \checkmark&  \checkmark & & 82.9 & 85.8\\
\checkmark & \checkmark&  \checkmark & \checkmark & 82.7 & 86.3\\
\checkmark & \checkmark &   & \checkmark & \textbf{87.9} & \textbf{92.0}\\
\end{tabular}
}
 \end{center}
\vspace{-0.4cm}
\caption{Ablation studies in the context-motion branch on the WF Pseudo Anomaly dataset. \textbf{GA} is global alignment; \textbf{VQ} is vector quantization; \textbf{LB} is local batch-wise alignment; \textbf{LP} is local patch-wise alignment; \textbf{AUC$_{mot}$} is reported only using a single branch; \textbf{AUC} is the final pseudo anomaly score.   }
\label{tab:WF_ablation}
\vspace{-0.3cm}
\end{table}
% \rnote{Did not edit.  Some wasted space with these tables.  How important are the ablations?}

% \begin{table}[ht]
% \centering
% \caption{Ablation Studies for Motion and Model 2}
% \label{table:ablation}
% \begin{tabular}{cccc|cc|ccc}
% \toprule
%  \multicolumn{4}{c}{\textbf{Cxt-Mot }} & \multicolumn{2}{c}{\textbf{Cxt-App }} & \multicolumn{3}{c}{\textbf{AUC}} \\
% \cmidrule(lr){1-4} \cmidrule(lr){5-6} \cmidrule(lr){7-9}
% \textbf{GA} & \textbf{VQ} & \textbf{LB} & \textbf{LP} & \textbf{GA}& \textbf{LP} & \textbf{Mot}& \textbf{App} &\textbf{+} \\
% \midrule
%  \checkmark & - & - & -  &\checkmark &- & 83.76 &76.45 & 89.33 \\
%  \checkmark&\checkmark& - & - & \checkmark & - & 86.44 &76.45 & 91.2 \\
%   \checkmark&\checkmark& \checkmark & - & \checkmark & - & 82.89 &76.45 & 91.2 \\
%  & - & X & - & - \\
% \midrule
%  - & - & - & Y & - \\
%  - & - & - & - & Y \\

% \bottomrule
% \end{tabular}
% \end{table}

\subsection{Detecting context-free anomalies}

% \begin{table}[!th]
%  \begin{center}
%     {
% \begin{tabular}{|c c c |}
% \toprule
%  \textbf{model} &\textbf{Ped2} & \textbf{Biker Day} \\    
% \midrule
%  \textbf{U-net}  & 94.3 & 92.4\\
%  \textbf{Trinity} & 97.9 &98.5\\
% \bottomrule
% \end{tabular}
% }
%  \end{center}
% \caption{Performance comparison on Bikerday data    }
% \label{tab:bikerday_auc}
% \end{table}

% A good example is the Test008 video shown in Fig.~\ref{fig:biker}, which contains a biker and a skater that are considered anomalies in the original setting. By changing the Bikerday context, the corresponding ground truth changes respectively. 
% Our model can learn to let bikers pass under the bikerday setting while reporting anomalies if we ban the biker as the original Ped2 dataset.

% We also notice there are several reverse error events where certain events like bus crossing only happen in certain contexts but are unseen in their paired context. This further suggests that Trinity captures the context-visual correspondence. There are several failure cases like enormous trucks with colorful advertisements entering the scenario or the flash headlights of firetrucks in the night. 

To apply Trinity to VAD problems in which there is no context information/dependence, we modify the loss functions in Section 3.2 by removing any term involving context (cxt).  Since there is no context information, the global term in Section 3.2.1 is replaced by $L_{global} = L^{global}_{app, mot}$ and computed similarly.

For inference, we use patch-wise local alignment between motion and appearance as the indicator for determining anomalies.  After computing the patch-wise local alignment logits $\vA^{patch}$ with the softmax function applied, a normal frame is expected to be similar to the identity matrix $\mathbf{I}$. Hence we use the Frobenius norm of the residual $\vS_{l} = \| softmax(\vA^{patch}) - \mathbf{I} \|_F $ as the anomaly indicator. We combine the next frame prediction quality from the U-net using PSNR and patch-wise local alignment as the final indicator of normalcy $\vS = \alpha \cdot \vS_{r}+(1-\alpha) \cdot \vS_{l}$.

Following the strategy described above, we evaluate Trinity without the context part on three traditional VAD benchmarks, and show that we can achieve comparable performance to state-of-the-art methods that are tailored to these benchmarks, as shown in Table \ref{tab:benchmark_auc}.  Again, we emphasize that Trinity was not designed for short, context-free datasets in the first place.  We provide these results to indicate that Trinity can also be applied successfully to VAD in isolated video clips.
\begin{table}[htbp!]
 \begin{center}
    {
\begin{tabular}{c|c|c|c}
\textbf{Methods}  & \textbf{Ped2}~\cite{weixin_li_anomaly_2014} & \textbf{Avenue}~\cite{lu_abnormal_2013} & \textbf{SHT}~\cite{liu_future_2018}   \\
\hline 
MemAE~\cite{gong_memorizing_2019} & 94.1 & 83.3 &71.2 \\MNAD~\cite{park_learning_2020}  & 97.0 &  88.5 &  72.8 \\
USTN-DSC~\cite{yang_video_2023}  & 98.1 & 89.9 & 73.8 \\
$HF^{2}$VAD~\cite{lv_learning_2021} & 99.3 & 91.1  &76.2   \\
DMAD~\cite{liu_diversity-measurable_2023}  & \textbf{99.7} &92.8  & 78.8  \\
HSC~\cite{sun_hierarchical_2023}  & 98.1 &\textbf{93.7}  & \textbf{83.4}  \\
\hline
\textbf{Ours}  & 97.9 & 88.5 & 74.1  
\end{tabular}
}
 \end{center}
 \vspace{-0.4cm}
\caption{Video anomaly detection results on benchmark datasets Ped2 \cite{weixin_li_anomaly_2014}, Avenue \cite{lu_abnormal_2013}, and ShanghaiTech (SHT) \cite{liu_future_2018}.  }
\label{tab:benchmark_auc}
 \vspace{-0.5cm}
\end{table}

\section{Conclusions and Future Work}
\label{sec:conclusion}
We proposed a context-aware video anomaly detection algorithm, introducing a new consideration for real-world VAD.  We carefully defined the task of detecting context-related anomalies, collected a novel dataset to enable our investigation, and proposed a contrastive learning framework to successfully detect context anomalies that cannot be found by traditional VAD methods. In future work, we plan to investigate a better negative sampling strategy for global alignment to avoid using negative samples that may have similar contexts.
% Further investigation of local level alignment to improve detecting non-context anomalies performance. 
% We proposed a context-aware video anomaly detection algorithm, introducing a new consideration for real-world VAD.  We carefully defined the task of detecting context-related anomalies, collected a novel dataset to enable our investigation, and proposed a contrastive learning framework to successfully detect context anomalies that cannot be found by traditional VAD methods. In future work, we plan to investigate a better negative sampling strategy for global alignment to avoid using negative samples that may have similar contexts. Further investigation of local level alignment for detecting non-context anomalies is also a promising direction. 
\section{Acknowledgment}
\label{sec:ack}
This material is based upon work supported by the U.S. Department of Homeland Security under Grant Award 22STESE00001-03-02. The views and conclusions contained in this document are those of the authors and should not be interpreted as necessarily representing the official policies, either expressed or implied, of the U.S. Department of Homeland Security.
{
    \small
    \bibliographystyle{ieeenat_fullname}
    \bibliography{main}
}

\clearpage
\setcounter{page}{1}
\maketitlesupplementary
\section{Architecture and training details}
\label{sec:Supp_archtecture}

\subsection{Architecture details}
\textbf{Context Branch.} \  \ In the context branch, the MLP consists of two `Linear-LayerNorm-GELU' blocks and one `Linear' block for projecting concatenated one-hot encoded context information into a 128-dimensional embedding. The number of heads in the Temporal/Spatial Context Attention Block (TSCAB) is set to 8.

\textbf{Motion Branch.}\ \ To calculate the background ratio of each non-overlapping patch for extracting HOF features, we set the flow magnitude thresholds to 0.5 and 1.0 for the benchmark datasets and the WF dataset respectively. We use $k_{mot} = 3$ transformer blocks to extract information from the quantized motion and the error code. For the motion transformer block, the number of heads is set to 8.

\textbf{Appearance Branch.}\ \ The spatial downsampling rate of the U-net is $8$ and the `Conv2d' embedding layer uses a size 2 kernel with stride 2 to embed $16\times 16$ spatial patches as tokens for the following transformer blocks.  We use $k_{app} = 3$ transformer blocks to extract appearance information.

All three branches generate output of the same size of $\vh \in \RR^{B\times N \times D}$, where the batch size $B$ is set to 12, the token length $N$ is set to 257 and 193 for the benchmark datasets and the WF dataset respectively, and the feature dimension $D$ is set to 512.

\subsection{Details about local level alignment}

\begin{figure}[!htbp]
\begin{center}
\includegraphics[width=0.99\linewidth]{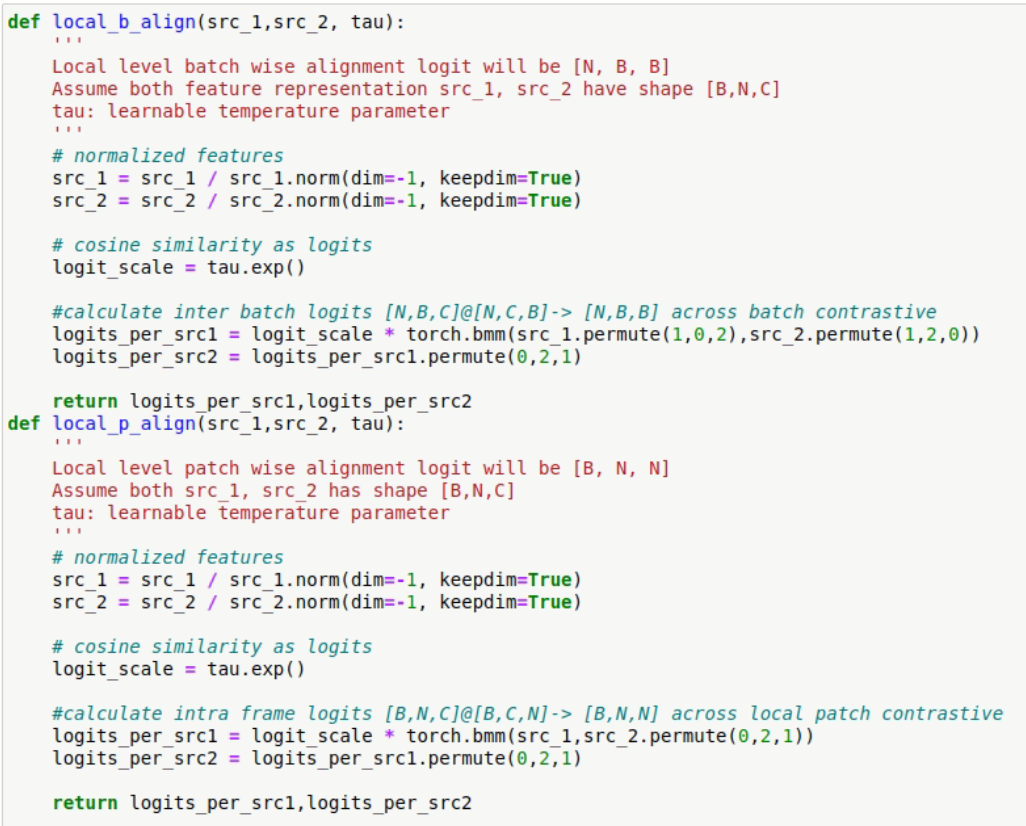}
\end{center}
\caption{Pytorch-style pseudo code for calculating local alignment logits. \textbf{Top:} the pseudo code of calculating batch-wise local alignment; \textbf{Bottom:} the pseudo code of calculating patch-wise local alignment. }
\label{fig:local_alignment}
\end{figure}

\begin{figure}[!htbp]
\begin{center}
\includegraphics[width=0.99\linewidth]{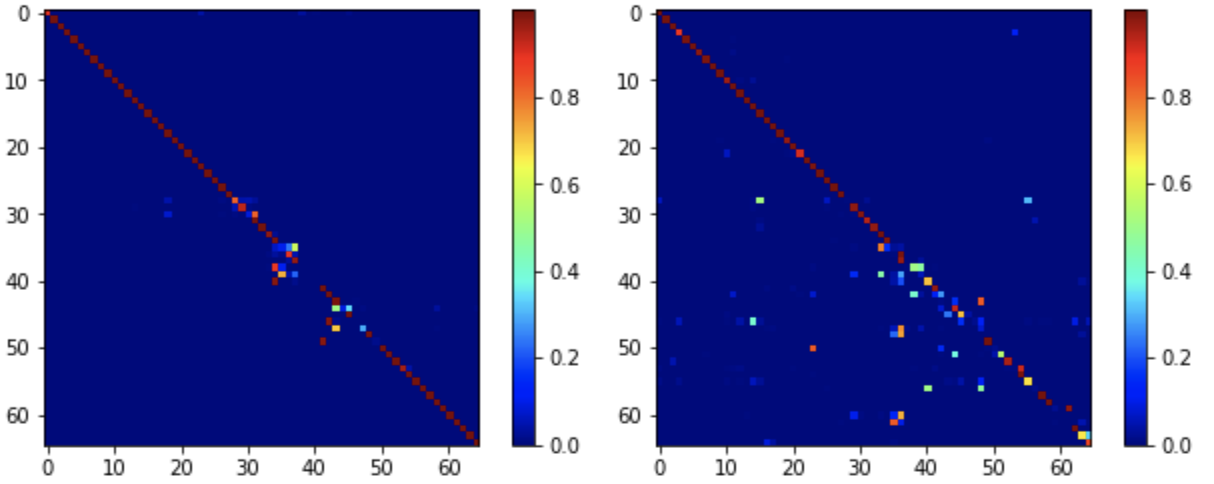}
\end{center}
\caption{Visualization of patch-wise local alignment results.  The x-y axis represents the spatial index of the token. \textbf{Left:} A normal frame result in the Ped2 dataset; \textbf{Right:} An abnormal frame result in the Ped2 dataset. }
\label{fig:patch_alignment}
\end{figure}

The pseudo-code for patch-wise local alignment is shown in Fig.~\ref{fig:local_alignment}. For patch-wise local alignment, the representation of each patch treats the rest of patches within the frame as negative samples. During inference, we utilize the patch-wise local alignment quality as the indicator of anomalies. As shown in Fig.~\ref{fig:patch_alignment}, a normal frame during inference will have an almost perfect diagonal matrix, indicating most patches have good alignment between the appearance and motion representations. On the contrary, when an anomaly appears in the frame, either unseen motion or unseen appearance will make the local alignment quality decrease drastically. Through evaluating the patch-wise local alignment quality, we can detect context-free anomalies.

\section{WF Dataset}
\label{sec:supp_data}

\begin{figure*}[th]
\includegraphics[width=0.99\linewidth]{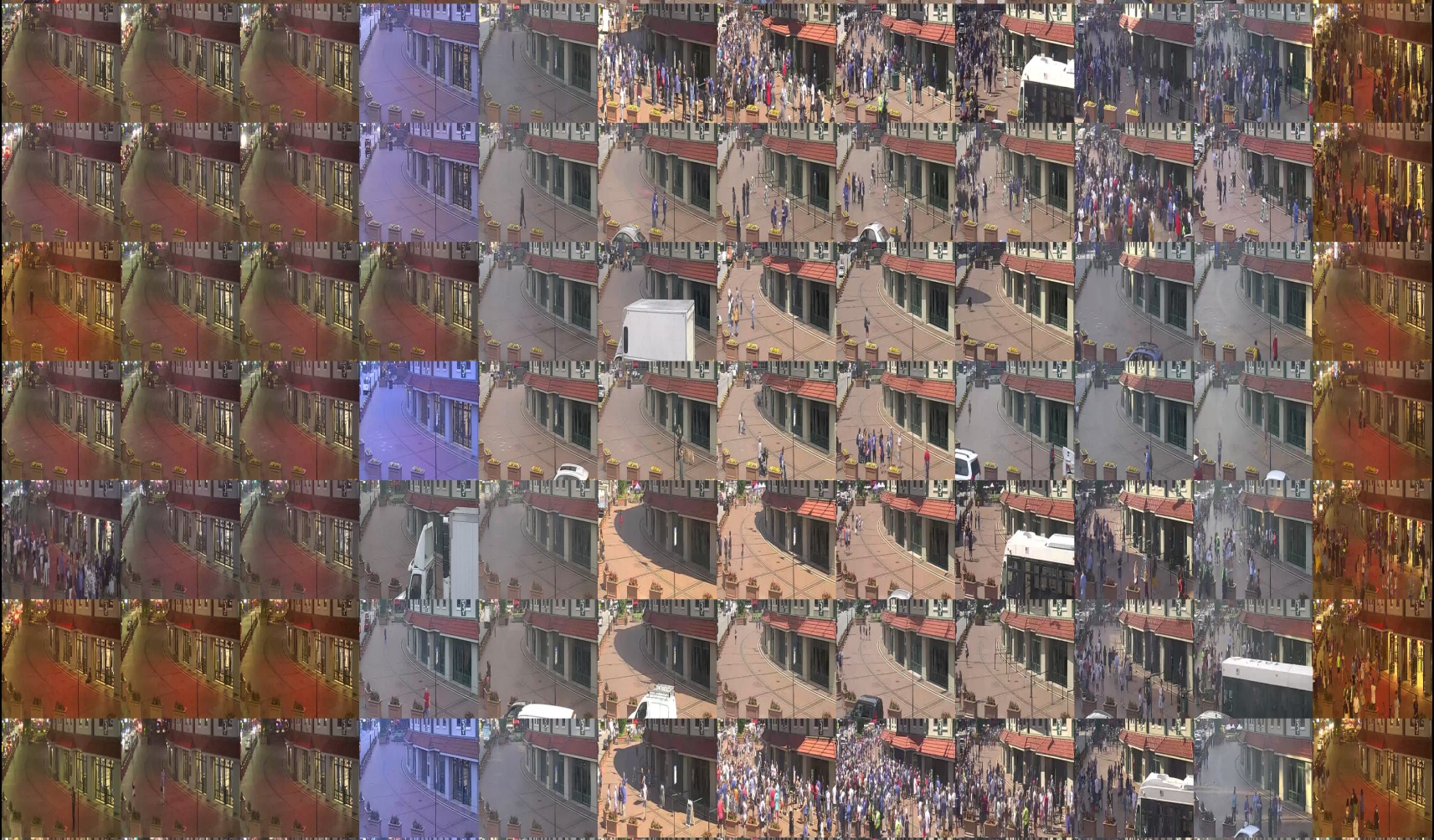}
\caption{Sample day grids from the WF Dataset. Each row represents a snapshot of a day 
 containing re-scaled frames that are evenly sampled every 2 hours from Wrigley Field Stadium (Best view in color).   }
\label{fig:day_grid}
\end{figure*}

In this work, the WF dataset used for investigating video context anomaly detection contains complex scenario dynamics. In Fig.~\ref{fig:day_grid}, we arrange video frames from the WF dataset in a grid format with each row showing 12 frames with a sampling gap of 2 hours within a day. The activities within the WF dataset vary significantly at different times of the day, and show strong correspondence with the game schedule. The context information for the video is organized as follows: The time of day is indicated on a 0 to 23 hour scale, and the day of the week is denoted by the numbers 0 to 6, representing Monday to Sunday. The game time indicator is set to 1 if the video's timestamp falls within 2 hours before and 3 hours after a game starts. The game schedule is marked similarly to the time of day, with a value of 0 indicating no game. Each context element is represented using one-hot encoding, resulting in a 56-dimensional binary vector through concatenation.

The pseudo-contextual anomalies that are introduced in \cref{sec:dataset} are achieved through altering the original context of normal videos with different intentions to mimic various context-anomalous scenarios. We provide examples of alteration from the pseudo-contextual anomalies in Table \ref{tab:wf_pseudo_annotation}. 

\begin{table*}
 \begin{center}
    {
\begin{tabular}{c |c|c  }
\textbf{Original Context}  & \textbf{Modification Intention }& \textbf{Altered Context Anomaly}   \\
\hline
10 4 0 \textbf{14} & day game morning $\rightarrow$ non-game day morning & 10 4 0 \textbf{0} \\
\textbf{16} 4 1 14  & during day game $\rightarrow$ time ahead to 2 hours earlier & \textbf{14} 4 1 14 \\
13 0 \textbf{0} \textbf{0}   & non-game day $\rightarrow$ game starts at 14:00 &  13 0 \textbf{1} \textbf{14}  \\
22 5 \textbf{1} \textbf{19} & night game people exiting the stadium $\rightarrow$ cancel game schedule & 22 5 \textbf{0} \textbf{0} \\
\hline
\end{tabular}
}
 \end{center}
\caption{Sample pseudo anomaly context alteration. The context information from the original context and altered context are listed as: time of day, day of the week, game indicator, game schedule respectively.}
\label{tab:wf_pseudo_annotation}
\end{table*}
% \section{Rationale}
% \label{sec:rationale}
% % 
% Having the supplementary compiled together with the main paper means that:
% % 
% \begin{itemize}
% \item The supplementary can back-reference sections of the main paper, for example, we can refer to \cref{sec:intro};
% \item The main paper can forward reference sub-sections within the supplementary explicitly (e.g. referring to a particular experiment); 
% \item When submitted to arXiv, the supplementary will already included at the end of the paper.
% \end{itemize}
% % 
% To split the supplementary pages from the main paper, you can use \href{https://support.apple.com/en-ca/guide/preview/prvw11793/mac#:~:text=Delete%20a%20page%20from%20a,or%20choose%20Edit%20%3E%20Delete).}{Preview (on macOS)}, \href{https://www.adobe.com/acrobat/how-to/delete-pages-from-pdf.html#:~:text=Choose%20%E2%80%9CTools%E2%80%9D%20%3E%20%E2%80%9COrganize,or%20pages%20from%20the%20file.}{Adobe Acrobat} (on all OSs), as well as \href{https://superuser.com/questions/517986/is-it-possible-to-delete-some-pages-of-a-pdf-document}{command line tools}.

\end{document}